\documentclass[runningheads]{llncs}
\usepackage[T1]{fontenc}
\usepackage{graphicx}
\usepackage{tikz}
\usetikzlibrary{shapes, arrows.meta, positioning}
\usepackage{float}

\newcommand\blfootnote[1]{%
  \begingroup
  \renewcommand\thefootnote{}\footnote{#1}%
  \addtocounter{footnote}{-1}%
  \endgroup
}

\begin{document}
\title{Impact of Feedback Type on Explanatory Interactive Learning}
%
%
\author{Misgina Tsighe Hagos\inst{1,3}\orcidID{0000-0002-9318-9417} \and
Kathleen M. Curran\inst{1,2}\orcidID{0000-0003-0095-9337} \and
Brian Mac Namee\inst{1,3}\orcidID{0000-0003-2518-0274}}
\authorrunning{M. T Hagos et al.}
%
\institute{Science Foundation Ireland Centre for Research Training in Machine Learning\\ 
\email{misgina.hagos@ucdconnect.ie}\\
\and
School of Medicine, University College Dublin\\
\email{kathleen.curran@ucd.ie}\\
\and
School of Computer Science, University College Dublin, Ireland\\
\email{brian.macnamee@ucd.ie}}

%
\maketitle              
\begin{abstract}

Explanatory Interactive Learning (XIL) collects user feedback on visual model explanations to implement a Human-in-the-Loop (HITL) based interactive learning scenario. Different user feedback types will have different impacts on user experience and the cost associated with collecting feedback since different feedback types involve different levels of image annotation. Although XIL has been used to improve classification performance in multiple domains, the impact of different user feedback types on model performance and explanation accuracy is not well studied. To guide future XIL work we compare the effectiveness of two different user feedback types in image classification tasks: (1) instructing an algorithm to ignore certain spurious image features, and (2) instructing an algorithm to focus on certain valid image features. We use explanations from a Gradient-weighted Class Activation Mapping (GradCAM) based XIL model to support both feedback types. We show that identifying and annotating spurious image features that a model finds salient results in superior classification and explanation accuracy than user feedback that tells a model to focus on valid image features.

 

\end{abstract}

\section{Introduction}
\label{introduction}



\blfootnote{Accepted at the International Symposium on Methodologies for Intelligent Systems, ISMIS 2022.}The need to involve humans (or experts) in the model training process, referred to as Human-in-the-Loop Learning (HITL), has inspired research on interactive and active learning \cite{kim2015interactive}. Interactive Machine Learning (IML) \cite{fails2003interactive} is a type of machine learning that adds human interaction to the model training process, as opposed to typical machine learning which aims to use training algorithms alone. One example of IML is active learning \cite{cohn1996active,settles2012active} in which a human participates in the model training process by providing labels for unlabelled instances. 

Even though the training process of IML and active learning is interactive, it falls short in involving humans at a detailed level of interaction. Humans are usually only involved in instance class labelling---a relatively low level task. A recent trend in interactive learning, referred to as Explanatory Interactive Learning (XIL) \cite{teso2019explanatory}, proposes richer interaction with humans by accepting user feedback on visual explanations (motivated by recent advances in Explainable Artificial Intelligence (AI) \cite{kenny2021post}). Explainable AI is a research area that focuses on providing understandable interpretations of AI models, which are usually considered as a black-box, to end users. User feedback in the form of annotations can be collected on explanations and used for model and explanation refinement \cite{schramowski2020making,selvaraju2019taking}.




In classification tasks where the class labels are mutually exclusive, mutual relations between image regions can be ignored and the expected user feedback can be narrowed down to the object in an image and the confounding image region. Assuming correct classification of instances, the two most common types of user feedback solicited in XIL are: (1) \emph{Missing Region} feedback: in which users identify regions that the model is currently ignoring, but should be focused upon; and (2) \emph{Spurious Region} feedback: in which users annotate regions that the model is currently focusing on but should have been ignored since they represent spurious signals. This gives rise to an obvious, but as yet unaddressed question: \emph{which type of feedback is more effective in XIL scenarios?} While the XIL literature explores various rich feedback collection mechanisms for model training; analysis and comparison of different user feedback types is largely ignored. Although it might seem obvious to collect both types of feedback, user interaction is time consuming and expensive. Moreover, different to active learning in which the cost of labelling instances is largely uniform across unlabelled instances \cite{settles2012active}, different feedback types in XIL have different impacts on user experience and the cost associated with soliciting feedback, since the expected feedback involves annotating image features which are not always uniform across input images. Given that hundreds or thousands of images are usually required for model training, the process of feedback collection in the form of annotation can take hours or days. Another aspect of XIL that is usually ignored in the literature is reporting its impact on the explanation localization accuracy of models. Since the basic idea behind XIL is using model explanations as a medium of interaction with annotators and a way to identify if a model is focusing on spurious features or if it is ignoring important regions of an image, the model training that follows should have a positive impact on the accuracy of the explanations too. In short, a model trained using XIL should be better at focusing on important regions and ignoring wrong image regions than models trained without feedback. 



In this paper, we compare the effectiveness of the two most common types of user feedback used in XIL on model performance and explanation accuracy. To do this we use Gradient-weighted Class Activation Mapping (Grad-CAM) \cite{selvaraju2017grad} as a feature attribution based model explanation to visualize salient regions of images. We train models on two decoy versions of the Fashion MNIST (FMNIST) dataset \cite{xiao2017fashion} using cross entropy classification loss and explanation loss that is computed between missing and spurious region feedback and GradCAM explanations. Furthermore, we compare the impact of using classification loss only and adding feedback to model training on model performance and explanation localization accuracy. The main contributions of this paper are:
\begin{enumerate}
    \item The first comparison in the XIL literature of the impact of different user feedback types on the performance of XIL algorithms;
    \item A class-wise decoy version of the FMNIST dataset is created and will be provided as a resource for future XIL research;
    \item Our experiments demonstrate that collecting spurious region feedback is more valuable than collecting missing region feedback in XIL.
\end{enumerate}


    


\section{Related Work}
\label{related_work}

\begin{figure}[ht] 
\vskip 0.2in
\begin{center}

    \begin{tikzpicture}[
        node distance=4ex and 0em,
        block/.style={rectangle, draw, fill=white!20, 
    text width=6.5em, text centered, rounded corners, minimum height=3em},
        line/.style={draw, -latex},
        ]

        \node [block] (1) {Model\\Training};
        \node [block, below right= of 1] (2) {Model\\Explanation};
        \node [block, below left= of 1] (3) {Feedback\\Collection};

        \path [line] (1.east) to[out=0, in=90] (2);
        \path [line] (2) to[out=-180, in=0] (3.east);
        \path [line] (3) to[out=90, in=180] (1);
    \end{tikzpicture}

\caption{The Explanatory Interactive Learning (XIL) loop.}
\label{fig:XILloop}
\end{center}
\vskip -0.2in
\end{figure}
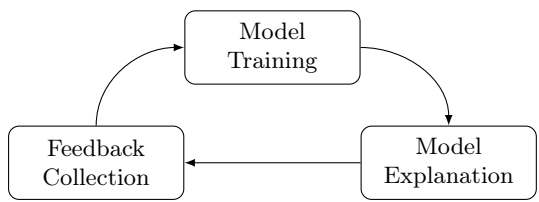

As shown in Fig. \ref{fig:XILloop}, XIL methods can be categorized based on the approaches they use during model training, model explanation, and user feedback collection. We explore existing work within XIL through these three lenses. 

\subsection{Model Training}

The two most common approaches to model training in XIL are model retraining and model fine-tuning.

\begin{itemize}
    \item Model-retraining. This approach utilises user feedback to produce new training examples, or counter examples, to mitigate against the impact of spurious features \cite{schramowski2020making}. In this approach, an expert is presented with explanations of a model's output showing features that a model finds useful for its prediction. Based on the provided explanations the expert provides feedback on whether the features are valid or not. Counter-examples are then fabricated by removing spurious features and adding valid features into a sample dataset for model refinement.
    \item Model fine-tuning (explanation losses.) In this category, an explanation loss penalty is added to the cost function (See Equation \ref{eq:ross_classification_loss_2017}) that is optimised during model training in addition to the loss terms that penalize a model for wrong predictions. This approach is summarised in Equations \ref{eq:schramowski} and \ref{eq:class_exp_loss} using GradCAM explanations, where \begin{math}M_{n} \in \ \{0,\ 1\}\end{math} is the ground truth annotation and \begin{math} norm \end{math} normalizes the Grad-CAM output, \begin{math} \theta \end{math} holds a model's parameters, with input \begin{math} X \end{math}, labels \begin{math}y \end{math}, predictions \begin{math}\hat{y} \end{math}, and a parameter regularization term \begin{math} \lambda \end{math}. Techniques such as Right for Right Reasons using Integrated Gradients (RRR-IG) \cite{ross2017right}, Right for the Right Reasons using GradCAM (RRR-GC) \cite{schramowski2020making}, and Right for Better Reasons (RBR) \cite{shao2021right} modify a model through explanation and training losses. Explanation losses can be computed between a feature annotations ground truth dataset and model generated explanations as can be seen in Equation \ref{eq:schramowski} \cite{schramowski2020making}.

\end{itemize}

\begin{equation}
    \label{eq:ross_classification_loss_2017}
    Classification\ Loss\ =\sum _{n=1}^{N}\sum _{k=1}^{K} -y_{nk}\log\hat{y}_{nk} 
\end{equation}


\begin{equation}
    \label{eq:schramowski}
    Explanation\ Loss =\ \sum _{n=1}^{N}( M_{n} * \ norm( GradCAM_{\theta }( X_{n})))^{2}
\end{equation}

\begin{equation}
    \label{eq:class_exp_loss}
    Loss\ =\ Classification\ Loss\ + Explanation\ Loss\ +\ \lambda\sum _{i} \theta _{i}^{2}
\end{equation}
    



\subsection{Model Explanation}

There are two major categories of model explanations used in XIL: (1) Local explanations that explain a single model outcome \cite{teso2019explanatory}; and (2) Surrogate model based learning, which uses a simple interpretable model to explain a more sophisticated black-box model, and to drive interaction with users \cite{popordanoska2020machine}. While surrogate models are effective for understanding the overall behaviour of a model, they may miss unique features that can be observed if local explanations are used.

\subsection{Feedback Collection}

Better feedback collection mechanisms will increase user involvement in XIL. Feedback in non-image domains, for example Recommender Systems (RS) and Natural Language Processing (NLP), can be more transparent since their explanations can be presented in a conversational natural language format and users can provide feedback using template questionnaires \cite{rago2021argumentative,madaan2021improving}---for example, Dalvi et al. (2022) used users' textual feedback on explanations to refine a model trained on a multiple choice questions dataset \cite{dalvi2022towards}. Models that learn and predict concepts \cite{koh2020concept} can simplify the feedback collection process because feedback is expected to be one of the learned concepts and a user only needs to detect if a wrong concept is being used for classification.


Due to the complex nature of image data, it is often more effective to use visual feedback mechanisms. Missing region and spurious region feedback are the two most commonly used types of user feedback in image-based XIL under the assumption of correct classification of instances. While techniques such as RRR-IG \cite{ross2017right}, RRR-GC \cite{schramowski2020making} and RBR \cite{shao2021right} use spurious region feedback to fine-tune a model to ignore spurious features, Human Importance-aware Network Tuning (HINT) trains a model to focus on valid image objects \cite{selvaraju2019taking}.

Although model explanations have been used for rich user interaction, the most effective types of user feedback that lead to high performance in XIL remain unknown. We compare the effectiveness of missing region and spurious region feedback, in terms of model performance and explanation accuracy, using two decoy versions of the FMNIST dataset. This has a potential to set the standard for future designs of explanation-based interactive machine learning.

\section{Methods}
\label{methods}

This section describes the datasets used, the experimental setup, and the model training process employed in the experiments described in this paper. 

\subsection{Dataset for XIL}

\begin{figure}[t]
    \vskip 0.2in
    \begin{center}
    \includegraphics[height=1.2cm]{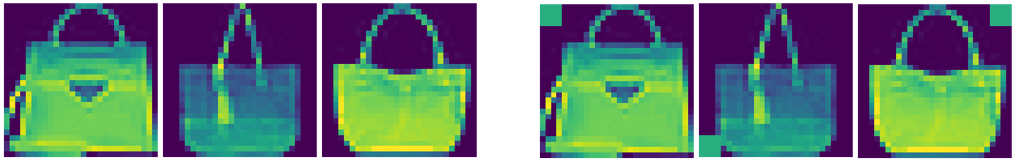}
    \vspace{-2mm}
    \caption{Sample images of class \emph{bag} from FMNIST (left) and random decoy FMNIST (right). Intensity values of pixels at the top right corner of the image from decoy FMNIST are exaggerated for presentation purposes.}
    \label{fig:example_decoy_fashionmnist}
    \end{center}
    \vskip -0.2in
    \vspace{-3mm}
\end{figure}

To demonstrate the effectiveness of XIL to refine models, Teso and Kersting \cite{teso2019explanatory} used a decoy version of the FMNIST dataset \cite{xiao2017fashion}. The decoy FMNIST dataset is made up of FMNIST images with 4x4 squares of high-intensity pixels added to randomly selected image corners (example images are shown in Fig.~\ref{fig:example_decoy_fashionmnist}). Locations of these confounders are class independent. In our work, in addition to experimenting with the decoy FMNIST dataset \cite{teso2019explanatory}, we also created a class dependent decoy version. Our decoy version of FMNIST contains confounders in the same image region across the same class. For easier reference, we refer to the Teso and Kersting \cite{teso2019explanatory} version as Random Decoy FMNIST, and our version as Class-wise Decoy FMNIST.

\subsection{Experimental Setup}
\label{section:experimental_setup}

Each of the two datasets contain 60,000 28x28 pixel images for model training. A test dataset of 10,000 unseen images each containing spurious region and object ground truth annotations is used to measure model performance. The spurious and missing region feedback are image masks of the added confounding regions and objects in images, respectively. We used accuracy to measure classification performance, and dice scores to assess explanation accuracy between model explanations and ground truth annotations.

\subsection{Model Training}

We trained six different models using the two datasets with the Adam optimizer and a decaying learning rate starting with \begin{math}1e^{-3}\end{math}. Two models are trained using only classification loss (cross entropy loss) using the Random and Class-wise Decoy FMNIST datasets. Two other models are trained using a combination of classification loss and explanation loss based on missing region feedback using each of the two datasets. The last two models are trained using a combination of classification loss and explanation loss based on spurious region feedback on the two datasets. After experimenting with different architectures and comparing their performances, a convolutional neural network with 3 convolutional layers followed by two fully connected layers was selected. We used a Right for the Right Reasons using GradCAM (Equation \ref{eq:schramowski}) explanation loss that is computed between GradCAM model explanations and missing region or spurious region feedback and L2 regularization to address overfitting.

\section{Results}
\label{results}

This section describes the results of the experiments performed, first exploring those based on the Random Decoy FMNIST dataset and then those based on the Class-wise Decoy FMNIST dataset.

\begin{table}[t]
\centering
\caption{Summary of feedback type influence on model performance.}
\label{table:results_summary}
\begin{tabular}{llrrrrrr} 
\hline
\multicolumn{2}{l}{~}                                                                                                     & \multicolumn{3}{c}{Random decoy FMNIST}                                                                                                        & \multicolumn{3}{c}{Class-wise decoy FMNIST}                                                                                                       \\ 

\multicolumn{2}{l}{Feedback type}                                                                                         & ~ None               & \begin{tabular}[c]{@{}c@{}}~ Missing\\~ Region\end{tabular} & \begin{tabular}[c]{@{}c@{}}~Spurious\\Region\end{tabular} & ~ None               & \begin{tabular}[c]{@{}c@{}}~ Missing\\~ Region\end{tabular} & \begin{tabular}[c]{@{}c@{}}~Spurious\\~Region\end{tabular}  \\ 
\hline
\multicolumn{2}{l}{\begin{tabular}[c]{@{}l@{}}Explanation accuracy\\against object annotations\end{tabular}}              & 0.27                 & 0.44                                                        & 0.65                                                      & 0.21                 & 0.44                                                        & 0.70                                                        \\ 

\multicolumn{2}{l}{\begin{tabular}[c]{@{}l@{}}Explanation accuracy\\ against spurious\\region annotations\end{tabular}}   & 0.05                 & 0.03                                                        & 0.02                                                      & 0.04                 & 0.03                                                        & 0.04                                                        \\ 

\multicolumn{2}{l}{\begin{tabular}[c]{@{}l@{}}Classification\\ accuracy\end{tabular}}                                     & 87.65                & 85.20                                                       & 85.53                                                     & 88.00                & 84.50                                                       & 86.21                                                       \\ 
\hline
 &                                                                                                                         & \multicolumn{1}{l}{} & \multicolumn{1}{l}{}                                        & \multicolumn{1}{l}{}                                      & \multicolumn{1}{l}{} & \multicolumn{1}{l}{}                                        & \multicolumn{1}{l}{}                                       
\end{tabular}
\vspace{-4mm}
\end{table}

\subsection{Random Decoy Fashion MNIST}
\label{results:class_inpdnt_decoy}

\subsubsection{Explanation Localization Accuracy}

\begin{figure}[t]
    \begin{center}
    \includegraphics[height=3.5cm,width=0.9\textwidth]{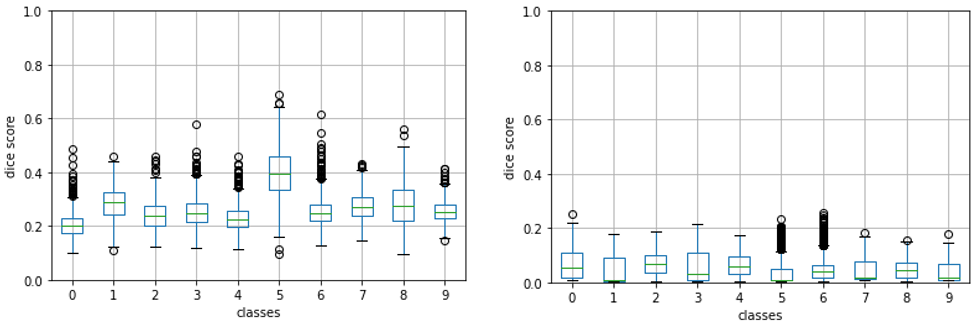}
    \caption{Dice score evaluation of a model trained on the random decoy dataset with classification loss only using object (left) and spurious region (right) ground truth annotations.}
    \label{fig:independent_class_loss}
    \end{center}
    \vskip -0.2in
\end{figure}

\begin{figure}[t]
    \begin{center}
    \includegraphics[height=3.5cm,width=0.9\textwidth]{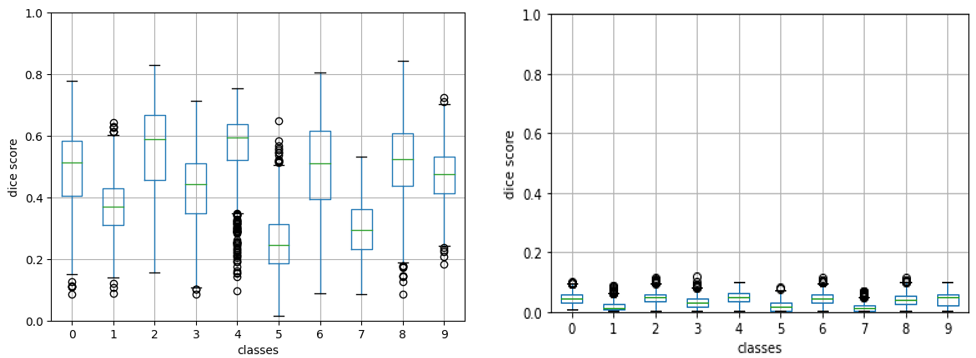}
    \caption{Dice score evaluation of a model trained on the random decoy dataset with classification loss and missing region feedback explanation loss using object (left) and spurious region (right) ground truth annotations.}
    \label{fig:independent_tr}
    \end{center}
    \vskip -0.2in
    \vspace{-3mm}
\end{figure}

\begin{figure}[t]
    \begin{center}
    \includegraphics[height=3.5cm,width=0.9\textwidth]{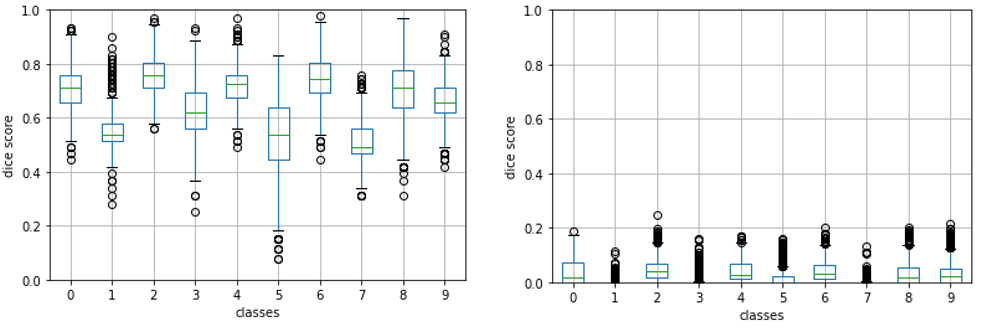}
    \caption{Dice score evaluation of a model trained on random decoy with classification loss and spurious region feedback explanation loss using object (left) and spurious region (right) ground truth annotations.}
    \label{fig:independent_wr}
    \end{center}
    \vskip -0.2in
\end{figure}

Figures \ref{fig:independent_class_loss}, \ref{fig:independent_tr} and \ref{fig:independent_wr} compare the performance of models trained using classification loss alone, and models trained with explanation losses using spurious region and missing region feedback on models' explanation accuracy using dice scores. Our target is to maximize dice scores that are computed against object annotations towards one, and minimize dice scores computed against spurious region towards zero. Average dice scores of explanations of the models trained on random decoy FMNIST compared against the test dataset of spurious region and object ground truth annotations is shown in Table~\ref{table:results_summary}. In both cases, the model which used spurious region feedback to compute explanation losses for training achieved superior performance.


\subsubsection{Classification Accuracy}

A summary of classification performance of the models trained on the random decoy dataset is displayed in Table~\ref{table:results_summary}. There is a slight performance loss in models that added feedback to their training compared to the model that only used classification loss. Compared against state of the art XIL methods, RRR-IG achieved the highest accuracy of 89.40, while RBR, HINT and RRR-GC scored 87.60, 58.20 and 78.60, respectively. All our models that used GradCAM explanations to compute explanation losses achieved superior performance scores compared to RRR-GC which also used GradCAM for model training.



\subsection{Class-wise Decoy Fashion MNIST}

\subsubsection{Explanation Localization Accuracy}

\begin{figure}[t]
    \begin{center}
    \includegraphics[height=3.5cm,width=0.9\textwidth]{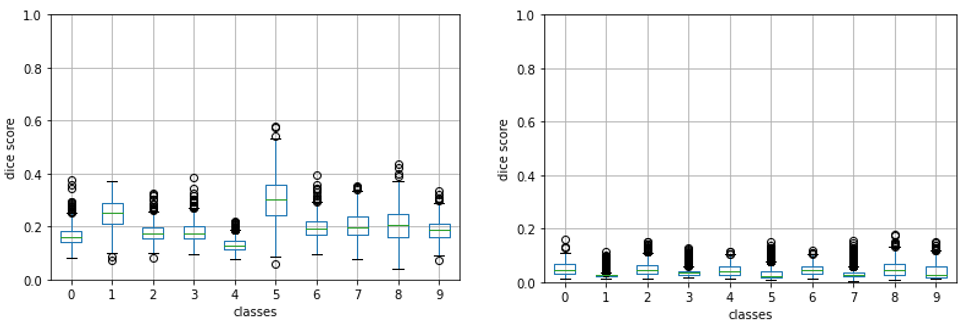}
    \caption{Dice score evaluation of a model trained on the class-wise decoy dataset with only classification loss using object (left) and spurious region (right) ground truth annotations.}
    \label{fig:dependent_class_loss}
    \end{center}
    \vskip -0.3in
\end{figure}

\begin{figure}[t]
    \begin{center}
    \includegraphics[height=3.5cm,width=0.9\textwidth]{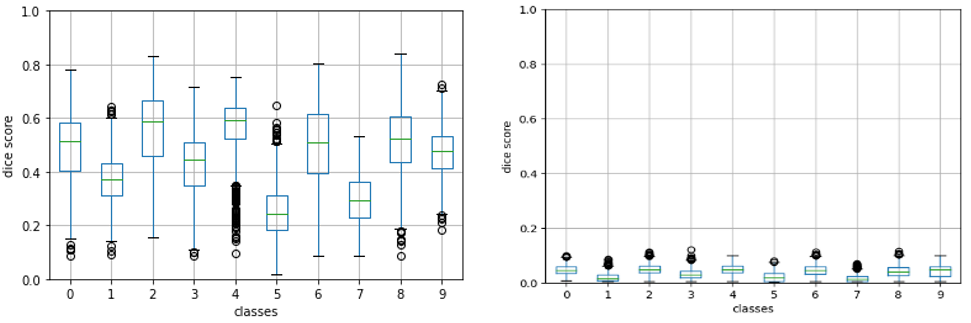}
    \caption{Dice score evaluation of a model trained on the class-wise decoy dataset with classification loss and missing region feedback explanation loss using object (left) and spurious region (right) ground truth annotations.}
    \label{fig:dependent_tr}
    \end{center}
    \vskip -0.2in
\end{figure}

\begin{figure}[t]
    \begin{center}
    \includegraphics[height=3.5cm,width=0.9\textwidth]{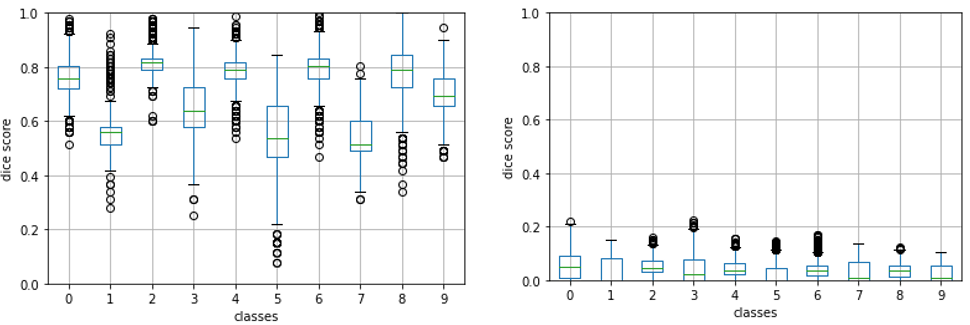}
    \caption{Dice score evaluation of a model trained on class-wise decoy with classification loss and spurious region feedback explanation loss using object (left) and spurious region (right) ground truth annotations.}
    \label{fig:dependent_wr}
    \end{center}
    \vskip -0.4in
\end{figure}

A dice score comparison between training with classification loss only, and training with added explanation losses using spurious region and missing region feedback on a model's explanation accuracy on the Class-wise Decoy FMNIST data is displayed in Figures \ref{fig:dependent_class_loss}, \ref{fig:dependent_tr} and \ref{fig:dependent_wr}. Average dice scores for explanations of models trained on the Class-wise Decoy FMNIST dataset compared against the test dataset of spurious region and object ground truth annotations is shown in Table~\ref{table:results_summary}. While the model trained with spurious region feedback achieved the highest average dice score when compared against object ground truth annotations, the model trained with missing region feedback achieved the best against spurious region annotations.
\vspace{-4mm}

\subsubsection{Classification Accuracy}


Average accuracy scores for all experiments are shown in Table~\ref{table:results_summary}. Except for the model that was trained using missing region feedback, the remaining two models achieved superior performance compared to their counterpart models that were trained on the random decoy FMNIST. 



\section{Conclusion}
\label{conclusion}

In this work, in addition to using a publicly available random decoy version of the FMNIST dataset, we generated a class-wise decoy version to compare effectiveness of different feedback types on a model's classification performance and explanation localization accuracy. We studied two feedback types: missing region and spurious region feedback. Apart from a slight classification performance loss when compared to a model trained using classification loss only, we achieved improved classification performance when compared against RRR-GC method that used GradCAM as explanations. In general, the models that utilized explanation losses achieved better explanation localization accuracy than models that only used classification losses. We believe the significant gains in explanation localization accuracy performance outweighs the slight loss of classification performance that these models suffered. More importantly, we were able to observe that using spurious region feedback is more valuable method to increase a model's classification and explanation accuracy than using missing region feedback. We believe our approach can be extended to other cases and investing on collecting spurious region feedback instead of missing region feedback has the potential to improve models and reduce associated cost. Most of the models trained on the class-wise decoy FMNIST performed better than those that were trained on the random version. We accredit this to the class-wise location uniformity of the confounding regions added to the class-wise version. We are aware that the feedback used in our experiments are accurate annotations which can not always be expected in real world scenarios and that feedback quality can be affected by the employed user interface medium. For this reason, we recommend performing user studies involving different feedback collection tools and comparing their performances before selecting one. For future work, we plan to compare impact of different feedback types with a user study.
\vspace{-4mm}





%
%
%

\subsubsection{Acknowledgements}

This publication has emanated from research conducted with the financial support of Science Foundation Ireland under Grant number 18/CRT/6183. For the purpose of Open Access, the author has applied a CC BY public copyright licence to any Author Accepted Manuscript version arising from this submission.

\bibliographystyle{samplepaper}
\bibliography{mybibliography}

\begin{thebibliography}{10}
\providecommand{\url}[1]{\texttt{#1}}
\providecommand{\urlprefix}{URL }
\providecommand{\doi}[1]{https://doi.org/#1}

\bibitem{cohn1996active}
Cohn, D.A., Ghahramani, Z., Jordan, M.I.: Active learning with statistical
  models. Journal of artificial intelligence research  \textbf{4},  129--145
  (1996)

\bibitem{dalvi2022towards}
Dalvi, B., Tafjord, O., Clark, P.: Towards teachable reasoning systems. arXiv
  preprint arXiv:2204.13074  (2022)

\bibitem{fails2003interactive}
Fails, J.A., Olsen~Jr, D.R.: Interactive machine learning. In: Proceedings of
  the 8th international conference on Intelligent user interfaces. pp. 39--45
  (2003)

\bibitem{kenny2021post}
Kenny, E.M., Delaney, E.D., Greene, D., Keane, M.T.: Post-hoc explanation
  options for xai in deep learning: The insight centre for data analytics
  perspective. In: International Conference on Pattern Recognition. pp. 20--34.
  Springer (2021)

\bibitem{kim2015interactive}
Kim, B.: Interactive and interpretable machine learning models for human
  machine collaboration. Ph.D. thesis, Massachusetts Institute of Technology
  (2015)

\bibitem{koh2020concept}
Koh, P.W., Nguyen, T., Tang, Y.S., Mussmann, S., Pierson, E., Kim, B., Liang,
  P.: Concept bottleneck models. In: International Conference on Machine
  Learning. pp. 5338--5348. PMLR (2020)

\bibitem{madaan2021improving}
Madaan, A., Tandon, N., Rajagopal, D., Yang, Y., Clark, P., Sakaguchi, K.,
  Hovy, E.: Improving neural model performance through natural language
  feedback on their explanations. arXiv preprint arXiv:2104.08765  (2021)

\bibitem{popordanoska2020machine}
Popordanoska, T., Kumar, M., Teso, S.: Machine guides, human supervises:
  Interactive learning with global explanations. arXiv preprint
  arXiv:2009.09723  (2020)

\bibitem{rago2021argumentative}
Rago, A., Cocarascu, O., Bechlivanidis, C., Lagnado, D., Toni, F.:
  Argumentative explanations for interactive recommendations. Artificial
  Intelligence p. 103506 (2021)

\bibitem{ross2017right}
Ross, A.S., Hughes, M.C., Doshi-Velez, F.: Right for the right reasons:
  Training differentiable models by constraining their explanations. arXiv
  preprint arXiv:1703.03717  (2017)

\bibitem{schramowski2020making}
Schramowski, P., Stammer, W., Teso, S., Brugger, A., Herbert, F., Shao, X.,
  Luigs, H.G., Mahlein, A.K., Kersting, K.: Making deep neural networks right
  for the right scientific reasons by interacting with their explanations.
  Nature Machine Intelligence  \textbf{2}(8),  476--486 (2020)

\bibitem{selvaraju2017grad}
Selvaraju, R.R., Cogswell, M., Das, A., Vedantam, R., Parikh, D., Batra, D.:
  Grad-cam: Visual explanations from deep networks via gradient-based
  localization. In: Proceedings of the IEEE international conference on
  computer vision. pp. 618--626 (2017)

\bibitem{selvaraju2019taking}
Selvaraju, R.R., Lee, S., Shen, Y., Jin, H., Ghosh, S., Heck, L., Batra, D.,
  Parikh, D.: Taking a hint: Leveraging explanations to make vision and
  language models more grounded. In: Proceedings of the IEEE/CVF international
  conference on computer vision. pp. 2591--2600 (2019)

\bibitem{settles2012active}
Settles, B.: Active learning. Synthesis lectures on artificial intelligence and
  machine learning  \textbf{6}(1),  1--114 (2012)

\bibitem{shao2021right}
Shao, X., Skryagin, A., Schramowski, P., Stammer, W., Kersting, K.: Right for
  better reasons: Training differentiable models by constraining their
  influence function. In: Proceedings of Thirty-Fifth AAAI Conference on
  Artificial Intelligence (AAAI) (2021)

\bibitem{teso2019explanatory}
Teso, S., Kersting, K.: Explanatory interactive machine learning. In:
  Proceedings of the 2019 AAAI/ACM Conference on AI, Ethics, and Society. pp.
  239--245 (2019)

\bibitem{xiao2017fashion}
Xiao, H., Rasul, K., Vollgraf, R.: Fashion-mnist: a novel image dataset for
  benchmarking machine learning algorithms. arXiv preprint arXiv:1708.07747
  (2017)

\end{thebibliography}

%




\end{document}